\pdfoutput=1
\documentclass[11pt,a4paper]{article}
\usepackage[T1]{fontenc}
\usepackage[utf8]{inputenc}
\usepackage{lmodern}
\usepackage[margin=2.5cm]{geometry}
\usepackage{amsmath,amssymb}
\usepackage{booktabs}
\usepackage[hidelinks]{hyperref}

\newcommand{\accBase}{9.5}
\newcommand{\accLP}{67.4}
\newcommand{\accSort}{67.0}
\newcommand{\diffSortLP}{$-0.4$}
\newcommand{\ciSortLP}{$[-2.9,\,+2.1]$}

\title{Just sort it!\\[2pt] \large A note on the LP behind rank-one anti-loop edits}
\author{Matteo Fischetti\\ \small DEI, University of Padova, Italy}
\date{November 2026}

\begin{document}
\maketitle

\begin{abstract}
Weight-space programming removes the thinking loops of a small reasoning model by a rank-one edit
$W' = W + u v^\top$ of an MLP output matrix, where $u$ is a loop direction and $v$ is the solution of a
sparse Linear Program (LP) over the neuron activations. We investigate whether the LP is needed.
We replace it by a sort: we take the 8 neurons whose activation has the smallest coefficient of
variation, and give them equal shares of the push that the LP solution realizes. On Qwen3.5-0.8B the
sorted edit scores \accSort\% on 1000 problems (MATH-500 and GSM8K) against \accLP\% for the LP edit and
\accBase\% for the unedited model; the paired difference is \diffSortLP{} points, with 95\% confidence
interval \ciSortLP{}, inside a non-inferiority margin of 5 points declared before the runs. On three
further sampling seeds the difference is $-0.6$ points, interval $[-2.0,\,+0.8]$, and $-0.8$ points,
interval $[-2.3,\,+0.6]$, when the dose is set without the LP. The sort
does not solve the LP, but the neurons it shares with the LP solution carry most of its mean push. The
outcome is that, on this model, the edit acts as a nearly constant bias along $u$.
\end{abstract}

\section{The LP}

In AHA~\cite{aha} the edit is applied to the \texttt{down\_proj} matrix of two layers of
Qwen3.5-0.8B~\cite{qwen}. Vector $u$ comes from matched pairs of looping and productive continuations
of the same prefix. Vector $v$ comes from the LP of~\cite{barry}: let $h_t$ be the activation vector at
the input of \texttt{down\_proj} on token $t$; then
\begin{align*}
\min\ & \textstyle\sum_j |v_j| + C \sum_t s_t\\
\text{s.t.}\ & v^\top h_t \le -\delta + s_t && \text{for the looping tokens,}\\
& |v^\top h_t| \le \varepsilon + s_t && \text{for the productive tokens,}\\
& |v_j| \le 10,\quad s \ge 0,
\end{align*}
followed by a restriction to the 8 neurons with largest $|v_j|\,\mathrm{std}(h_j)$ and by a second
solve.

The LPs are not small, and they are dense. In our run of AHA on Qwen3.5-0.8B, the LP of each layer has
12\,000 constraint rows (4000 looping and 8000 productive tokens, the latter with a two-sided
constraint), 3584 variables $v_j$ (the width of the MLP) plus one slack per row, and a dense constraint
matrix with about 43 million nonzeros. It is solved twice per layer, on all neurons and then on the 8
selected ones, for each of the 20 layers on which AHA solves it. Note that the optimum of the first
solve is very sparse: between 6 and 189 nonzero $v_j$ out of 3584. With the GPU implementation of the
primal-dual hybrid gradient method (PDHG) shipped with AHA, the search of a layer, essentially the two
solves, takes 3 to 11 seconds, about 2.2 minutes over the 20 layers, on one NVIDIA RTX PRO 6000
Blackwell. By contrast, the sort of Section~2 needs two moments per neuron, i.e., one pass over the
activations of the fit tokens.

At first glance, this is a separation model: push the looping tokens, leave the productive ones
alone. However, the solutions do not separate. On our fit rows the LP edit pushes the looping tokens by
$-0.79$ and the productive ones by $-0.70$ at layer 15 ($-0.89$ and $-0.83$ at layer 19), with a
coefficient of variation of the push of $0.21$ ($0.12$). The public sweep of~\cite{barry} tells the
same story: multiplying $\delta$ by 2 or by 8 leaves the accuracy unchanged, whereas $\varepsilon=0.25$
destroys it. Hence the LP synthesizes a bias along $u$, by using neurons that are almost always active.

\section{The sort}

If a bias is all one needs, the best neurons are those whose activation is as constant as possible.
For each neuron $j$ we compute the mean $m_j$ and the standard deviation $\sigma_j$ of $h_j$ on the fit
rows, with the class weights of the LP. We call \texttt{sort\_k8} the edit defined by the $k=8$ neurons
with smallest $\sigma_j/|m_j|$ and by the weights $v_j := c/(k\,m_j)$, where the dose $c$ is the mean
push of the LP solution on the same rows. Note that the dose is borrowed from the LP on purpose, so
that the two edits differ in the neurons and not in the dose; the window of useful doses is known to be
narrow~\cite{barry}. As a control we also consider \texttt{knap\_k8}, that takes the 8 neurons with
largest $|m_j|$ (the continuous knapsack rule, variance ignored). No solve is involved: two moments
per neuron and a sort.

\section{Results}

We ran the unmodified AHA pipeline (commit \texttt{e6e94ff5}) on Qwen3.5-0.8B with layers 15 and 19
fixed, on a single NVIDIA RTX PRO 6000 Blackwell, with one sample per problem. The protocol, the arms
and the verdict rule were committed before any number was produced.

\begin{table}[ht]
\centering
\caption{Accuracy in \%, seed 0. ``Diff'' is the paired difference with respect to \texttt{lp} on the
1000 problems; the 95\% confidence interval comes from the discordant pairs.}
\label{tab:acc}
\begin{tabular}{lrrrrr}
\toprule
arm & MATH-500 & GSM8K test & pooled & diff & 95\% CI\\
\midrule
\texttt{base} & 6.6 & 12.4 & 9.5 & $-57.9$ & $[-61.0,\,-54.8]$\\
\texttt{lp} & 61.4 & 73.4 & 67.4 & & \\
\texttt{sort\_k8} & 59.8 & 74.2 & 67.0 & $-0.4$ & $[-2.9,\,+2.1]$\\
\texttt{knap\_k8} & 57.4 & 75.0 & 66.2 & $-1.2$ & $[-3.7,\,+1.3]$\\
\bottomrule
\end{tabular}
\end{table}

Table~\ref{tab:acc} reports the outcome on MATH-500~\cite{math500} and on the first 500 problems of the
GSM8K test set~\cite{gsm8k}. \texttt{sort\_k8} and \texttt{lp} disagree on 166 problems, 81 won by the
sort and 85 by the LP. The knapsack-like rule \texttt{knap\_k8} is also within the margin, and the
difference between the two sorts is not significant ($+0.8$ points for \texttt{sort\_k8}, interval
$[-1.6,\,+3.2]$). Hence, on the task, we cannot tell the coefficient of variation from the mean as a
sorting key; the structural measures of Table~\ref{tab:struct} do separate them.

\begin{table}[ht]
\centering
\caption{The three vectors $v$ on the fit rows. ``Loop'' and ``prod'' are the mean pushes $v^\top h$
on the two classes, ``CV'' is the coefficient of variation of the push, ``hit'' is the fraction of
looping rows with $v^\top h \le -\delta$, ``hinge'' is the mean LP slack per row, ``common'' counts the
neurons shared with the LP support (out of 8).}
\label{tab:struct}
\begin{tabular}{llrrrrrr}
\toprule
layer & arm & loop & prod & CV & hit & hinge & common\\
\midrule
15 & \texttt{lp} & $-0.789$ & $-0.701$ & 0.212 & 0.991 & 0.005 & \\
15 & \texttt{sort\_k8} & $-0.768$ & $-0.729$ & 0.148 & 0.974 & 0.024 & 3\\
15 & \texttt{knap\_k8} & $-0.950$ & $-0.659$ & 0.485 & 0.845 & 0.523 & 4\\
\midrule
19 & \texttt{lp} & $-0.890$ & $-0.834$ & 0.123 & 0.910 & 0.062 & \\
19 & \texttt{sort\_k8} & $-0.882$ & $-0.864$ & 0.159 & 0.763 & 0.319 & 5\\
19 & \texttt{knap\_k8} & $-1.062$ & $-0.763$ & 0.384 & 0.810 & 0.724 & 4\\
\bottomrule
\end{tabular}
\end{table}

As to the structure, the sort is not the LP optimum: its hinge value is about 5 times larger, and at
layer 19 it satisfies the loop constraint on 76\% of the rows against 91\%. Note, however, that the
neurons shared by the two supports carry 95\% (layer 15) and 88\% (layer 19) of the mean push
$\sum_j v_j m_j$ of the LP solution. The other LP neurons contribute little to the mean push, although
some of them have a large weight (neuron 77 at layer 19 has $|v_j| = 0.53$): they shape the hinge term,
not the dose. Interestingly, this correction does not show up in the accuracy.

\subsection*{Sampling seeds}

After seeing the outcome of seed 0 we declared a second protocol, again before the runs: the same
edits, rebuilt from the stored $u$ and $v$, evaluated with three new sampling seeds. Note that the
seeds measure the sampling noise on fixed edits; they do not replicate the pipeline that produces $u$
and the LP solution. The second protocol adds one arm, \texttt{sortd\_k8}, with the neurons and the
shares of \texttt{sort\_k8} but with dose $c := -\delta$, where $\delta$ is computed by AHA from $u$
before any LP is solved. Hence \texttt{sortd\_k8} takes nothing from the LP solution. For every problem
we average the correctness over the three seeds, and the standard error is computed over the 1000
problems.

\begin{table}[ht]
\centering
\caption{Seeds 1, 2 and 3. Accuracy in \%, averaged over the seeds. ``Diff'' is the paired
difference with respect to \texttt{lp}, with the 95\% confidence interval clustered by problem.}
\label{tab:seeds}
\begin{tabular}{lrrrrr}
\toprule
arm & MATH-500 & GSM8K test & pooled & diff & 95\% CI\\
\midrule
\texttt{lp} & 59.9 & 74.3 & 67.1 & & \\
\texttt{sort\_k8} & 60.2 & 72.7 & 66.5 & $-0.6$ & $[-2.0,\,+0.8]$\\
\texttt{sortd\_k8} & 57.9 & 74.7 & 66.3 & $-0.8$ & $[-2.3,\,+0.6]$\\
\texttt{knap\_k8} & 58.5 & 73.9 & 66.2 & $-0.9$ & $[-2.3,\,+0.5]$\\
\bottomrule
\end{tabular}
\end{table}

Table~\ref{tab:seeds} confirms the outcome of seed 0, with an interval about half as wide: both
\texttt{sort\_k8} and \texttt{sortd\_k8}, whose dose does not come from the LP, are well inside the
margin of 5 points. The two sorting keys remain indistinguishable: \texttt{sort\_k8} minus
\texttt{knap\_k8} is $+0.3$ points, interval $[-1.2,\,+1.8]$. In a comparison not declared in the
protocol, \texttt{sortd\_k8} splits by benchmark: on MATH-500 it is 2.1 points below \texttt{lp}
(interval $[-4.2,\,0.0]$) and 2.3 points below \texttt{sort\_k8} (interval $[-4.3,\,-0.3]$), whereas on
GSM8K it is above both. A smaller dose seems to cost on the harder problems.

We did not test a constant shift along $u$ at the same dose. On a development set of 128 problems, the
sweep of~\cite{barry} reports 57.8\% for such a steering against 62.5\% for the LP edit, a gap within
the noise of 128 problems but in the direction of the LP; hence ``nearly constant'' is the right word, and a steering arm is the
natural next control.

\section{Conclusions}

We have shown that, on Qwen3.5-0.8B, a sort of the neurons by coefficient of variation matches the LP
of weight-space programming over three sampling seeds (difference $-0.6$ points, interval
$[-2.0,\,+0.8]$), without any solver. The dose,
a single scalar per layer, can be taken from the threshold $\delta$ that AHA computes before the LP.
Admittedly, we tested one model and one run of the pipeline that produces $u$. In our view the informative part of the AHA edit is the
direction $u$, not the LP that implements it.

\section*{Role of AI}

This note was produced with an AI assistant, and we prefer to state the division of labor plainly.
The idea is the author's: the LP of~\cite{barry} looked like a knapsack in disguise, hence its
solution might be replaced by a sort. The author also supervised the work, decided what to run and
what to publish, and takes responsibility for the content. Everything else was done by Claude
(Anthropic; models Fable 5.2 and Opus 5.5, used within Claude Code): reading the material of AHA and
of~\cite{barry}, the choice of the sorting key and of the dose, the two protocols, the code, the runs
on the cluster, the statistical analysis, and the text of this note, including this section. The
whole work took one day, 16 November 2026.

\paragraph{Data and code.} The two pre-registered protocols, the code, the vectors $u$ and $v$ of all
arms and the per-problem results are available, with this note, at
\url{https://doi.org/10.5281/zenodo.24286137}.

\end{document}